\def\year{2020}\relax
\documentclass[letterpaper]{article} 
\usepackage{aaai20}  
\usepackage{times}  
\usepackage{helvet} 
\usepackage{courier}  
\usepackage[hyphens]{url}  
\usepackage{graphicx} 
\usepackage{graphicx}  
\newcommand{\mname}{\texttt{AdaCare}\xspace}

\usepackage{enumitem}
\usepackage{array}
\usepackage{multirow}
\usepackage{subfig}
\usepackage{scrextend}
\usepackage{xspace}
\usepackage{xcolor}
\usepackage{amsmath}
\usepackage{bm}
\usepackage{amsfonts}
\usepackage{amsthm}

\title{\mname: Explainable Clinical Health Status Representation Learning via Scale-Adaptive Feature Extraction and Recalibration}

\author{Liantao Ma\textsuperscript{\rm 1,\rm 3}, Junyi Gao\textsuperscript{\rm 1, \rm 2}, Yasha Wang \textsuperscript{\rm 1,\rm 2}\thanks{Corresponding Author}, Chaohe Zhang\textsuperscript{\rm 1,\rm 3}, Jiangtao Wang\textsuperscript{\rm 4}, \\ \Large \textbf{Wenjie Ruan}\textsuperscript{\rm 4}, \textbf{Wen Tang\textsuperscript{\rm 5},  Xin Gao\textsuperscript{\rm 1,\rm 3}, Xinyu Ma\textsuperscript{\rm 1, \rm 3}}
\\ 
\textsuperscript{\rm 1}Key Laboratory of High Confidence Software Technologies, Ministry of Education, Beijing, China\\ 
\textsuperscript{\rm 2}National Engineering Research Center of Software Engineering, Peking University, Beijing, China\\ 
\textsuperscript{\rm 3}School of Electronics Engineering and Computer Science, Peking University, Beijing, China\\ 
\textsuperscript{\rm 4}School of Computing and Communications, Lancaster University, UK\\ 
\textsuperscript{\rm 5}Division of Nephrology, Peking University Third Hospital, Beijing, China    
\\
\{malt, wangyasha\}@pku.edu.cn, \{jiangtao.wang, wenjie.ruan\}@lancaster.ac.uk, tanggwen@126.com
}

\begin{document}

\maketitle

\begin{abstract}

Deep learning-based health status representation learning and clinical prediction have raised much research interest in recent years.
Existing models have shown superior performance, but
there are still several major issues that have not been fully taken into consideration.
First, 
the historical variation pattern of the biomarker in diverse time scales plays a vital role in indicating the health status, but it has not been explicitly extracted by existing works.
Second, key factors that strongly indicate the health risk are different among patients.
It is still challenging to adaptively make use of the features for patients in diverse conditions.
Third, using prediction models as the black box will limit the reliability in clinical practice. 
However, none of the existing works can provide satisfying interpretability and meanwhile achieve high prediction performance.
In this work, we develop a general health status representation learning model, named \mname. 
It can capture the long and short-term variations of biomarkers as clinical features to depict the health status in multiple time scales.
It also models the correlation between clinical features to enhance the ones which strongly indicate the health status and thus can maintain a state-of-the-art performance in terms of prediction accuracy while providing qualitative interpretability.
We conduct a health risk prediction experiment on two real-world datasets.
Experiment results indicate that \mname outperforms state-of-the-art approaches and provides effective interpretability, which is verifiable by clinical experts.

\end{abstract}

\noindent

\section{Introduction}


Health status prediction (e.g., mortality risk prediction, disease prediction) is of great interest to physicians.
For inpatients or patients with chronic diseases who face severe life threats and receive long-term treatments,
their health conditions are complex and continually changing over time.
By predicting patient's health status, physicians can select personalized follow-up treatments, prevent adverse outcomes, assign medical resources effectively, and reduce the medical cost.
Normally,
some biomarkers, such as blood albumin and blood glucose, are recorded through the treatment trajectories and further have been taken into consideration for the predication.
In a practical diagnosis process, physicians need to comprehensively evaluate the health of patients by identifying the high-risk factors. The precise risk prediction requires a high level of clinical expertise and experience.

\begin{figure}[]
  \centering
  \includegraphics[width=0.95\columnwidth]{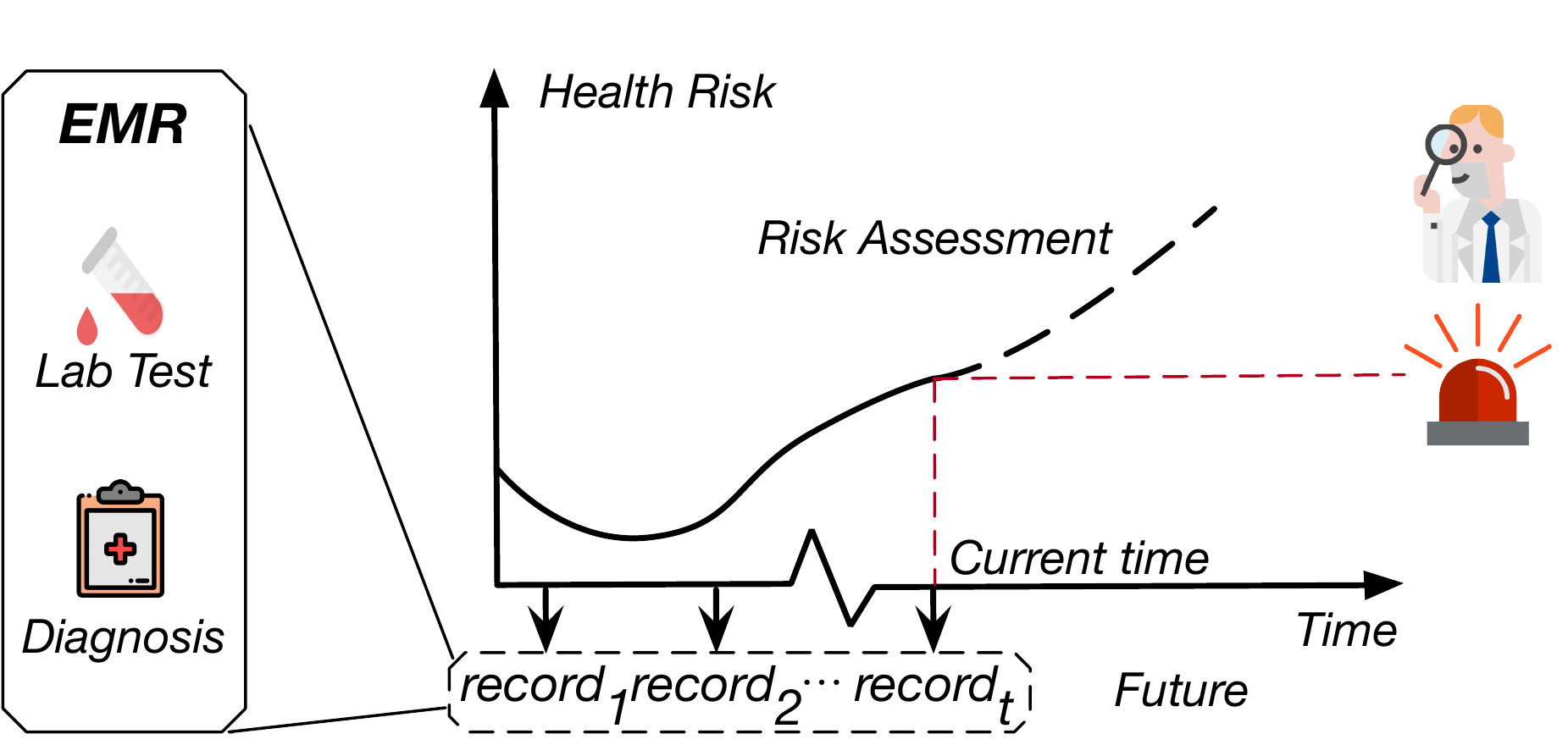}
  \caption{Perform dynamic patient health risk prediction to trigger the alarm for adverse outcomes and take an early individualized intervention.}
  \label{fig:introduction}
\end{figure}


Nowadays, electronic healthcare information systems are widely used in various healthcare institutions, which can precisely record the lab test results and health information of patients in terms of Electronic Medical Records (EMR). As depicted by Figure~\ref{fig:introduction}, EMR can be seen as a type of multivariate time series data and provide essential healthcare information for the data-driven healthcare prediction.

Recently, due to the remarkable representation learning ability of deep neural networks, many deep learning-based models have been developed to tackle such prediction tasks by using EMR data, including mortality prediction \cite{ma2020concare}, disease diagnosis prediction \cite{lee2018diagnosis}, and patient phenotype identification \cite{baytas2017patient}.
Usually, those models first embed the EMR data into low-dimensional feature space to learn the dense representation of the patients' health status and then perform specific clinical analysis tasks based on such representation. 
However, there are still some issues that are not yet fully resolved {by existing research works}, i.e., how to effectively embed temporal health information comprehensively, and how to assure the trustworthiness of the {representation learning} model in terms of providing verifiable interpretations. The issues are summarized as follows:

\begin{figure}[]
  \centering
  \includegraphics[width=0.95\columnwidth]{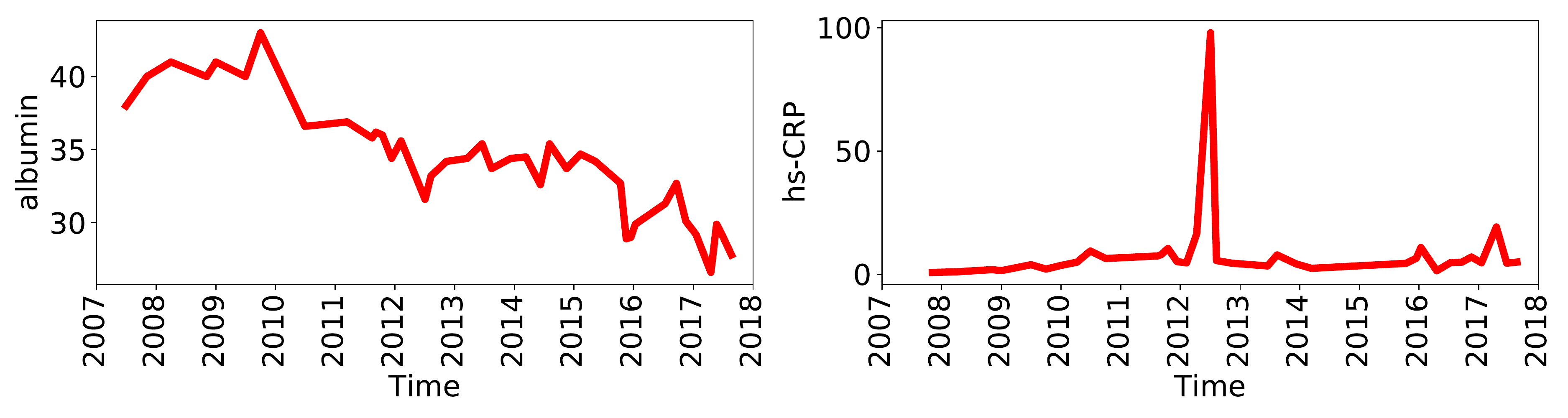}
  \caption{The long-term descending trend of blood albumin (left) and short-term abnormal rising of hypersensitive C-reactive protein (right) indicate the health risk.}
  \label{fig:lschange}
\end{figure}

\begin{itemize}[leftmargin=*]

\item 
\textbf{$I_1$: Historical variation pattern of the biomarker in different time scales}. 
Not only the value of lab test results but also its variation pattern contains essential information on the health status.
The long-term variation trend and short-term abnormal variation of different biomarkers both show the health status of patients from different aspects. 
Thus, it is vital that the variation of different biomarkers can be captured at different time scales.
For example, for end-stage renal disease (ESRD) patients, the long-term descending trend of blood albumin is a strong indicator of malnutrition and health deterioration~\cite{bharadwaj2016malnutrition}.
On the contrary, the short-term abnormal rising of hypersensitive C-reactive protein indicates the high-risk clinical event (e.g., peritonitis)~\cite{Ma2018Predictors}. 
Although several research works try to utilize the convolutional operator to extract temporal patterns of clinical events~\cite{cheng2016risk,ma2018health}, none of them can capture such patterns in multi-level time scales simultaneously.

\item { \textbf{$I_2$: Adaptively making use of the clinical features for patients in diverse conditions}.
Key factors that strongly indicate the health risk are different among patients ~\cite{valko2010feature}. 
So it is critical that the model can utilize the features adaptively when learning the health status representation and performing prediction for patients in different conditions. 
The model adaptability can be specifically expressed in two aspects:
\begin{itemize}
    \item The importance of features in different time-scales varies among different patients. For example, for the patient suffering from chronic disease, the feature extracted in the long term may be more representative for depicting the health status. On the contrary, for the patient diagnosed with acute disease, the short-term feature describes the health risk more precisely.
    
    \item The importance of features should also be adaptive to different characteristics. For example, the model should pay more attention to creatinine and urea when they rise \footnote{Plasma concentrations of creatinine and urea are usually associated with systemic manifestations (uremia) for chronic kidney disease patients \cite{msdmanuals_ckd}.}, and the model should concentrate more on diastolic blood pressure when the patient has been diagnosed with cerebrovascular disease\footnote{The level of diastolic blood pressure is usually associated with cerebrovascular disease \cite{rabkin1978predicting}}.
\end{itemize}

\item \textbf{$I_3$: Interpretability for various patients}.
Medical experts need to understand how a certain decision is made by a model to a particular patient at different visits.
So that the prediction results are trustworthy for developing individualized intervention and extracting medical knowledge~\cite{tangri2011determining}. 
For example, if the model triggers a health warning by taking the creatinine and urea as key factors, the physician is alerted to assess such patients for possible systemic manifestations.
Moreover, it will also remind the physicians of the previously unknown correlation between the biomarker and death reason.
However, most of the existing works can only provide visit-level or disease-feature-level interpretability by attention mechanism \cite{ma2017dipole,bai2018interpretable}. 
As far as we know, RETAIN~\cite{choi2016retain} is the only work that can provide reasonable biomarker-feature-level interpretability by utilizing two-level attention as an end-to-end model, but its prediction accuracy is unsatisfactory~\cite{ma2018health,ma2018risk}. To bridge this research gap, our model \mname can provide a fine-grained feature-level interpretation for the model prediction, but also achieves a state-of-the-art prediction accuracy.
}

\end{itemize}

By jointly considering the above research issues in clinical practice, we propose a clinical health status representation learning model via scale-\underline{ada}ptive feature extraction and recalibration, \underline{\mname}\footnote{We release our code and case studies at GitHub \url{https://github.com/Accountable-Machine-Intelligence/AdaCare}}.
It monitors biomarkers in long and short time scales simultaneously to extract temporal variation patterns, depicting the health status comprehensively for patients in diverse conditions (e.g., diagnosed with chronic diseases or acute diseases). 
\mname models the high relationship between clinical features to abstract the input.
At each visit, \mname selects the most indicative medical features to build health status representation.
Empirical studies show \mname boosts the performance and meanwhile offers key features that lead to the prediction.
Our main contributions are summarized as follows:

\begin{itemize}[leftmargin=*]
\item We build a general health status representation learning model, \mname, to effectively embed the health status and provide reasonable interpretability for patients in diverse conditions. \mname uses the dilated convolution with multi-scale receptive fields to capture the long and short-term variation patterns of biomarkers as clinical features and depict patient health status more comprehensively (addressing \textbf{$I_1$}).

\item We build the scale-adaptive feature recalibration module, which explicitly and adaptively models the feature relationship based on squeeze-and-excitation block \cite{hu2018squeeze} to selectively enhances high-risk features and meanwhile suppress the useless ones (addressing \textbf{$I_2$}).
And thus, \mname can provide interpretability on health status representation learning for patients in diverse health conditions as an end-to-end model, and further remind the physicians with the precursor of health risk (addressing \textbf{$I_3$}). 
Such interpretability is indicative of understanding how the model utilizes EMR data to make the assessment and extract valuable medical knowledge.

\item We conduct two prediction tasks (i.e., decompensation prediction and mortality prediction) on two real-world datasets (i.e., MIMIC-III dataset and end-stage renal disease dataset) respectively to verify the performance.
The results show that \mname outperforms the baseline approaches in both tasks.
The interpretability of \mname is demonstrated by an overall observation of feature recalibration.
Besides, the obtained medical knowledge has been positively confirmed by clinical experts.

\end{itemize}

\section{Related Work}

Over the past ten years, there has been a massive explosion in the amount of digital information stored in electronic medical records, which opens a door for researchers to make secondary use of these records for various clinical applications.
Deep learning-based models have shown the capability to perform
mortality prediction,
patients subtyping,
and
diagnosis prediction.
Though the medical tasks vary from each other,
their essences are usually learning the health status representations of patients.
There are two essential concerns among deep-learning-based EMR analysis researches:

\subsection{Temporal Medical Feature Extraction}
Some researches tried to extract high-level temporal clinical features by convolution modules as well as performing healthcare prediction \cite{cheng2016risk,ma2018health}.
However, according to medical experience, 
the variation of biomarkers should be evaluated in different time scales simultaneously when evaluating the patient's condition.
To the best of our knowledge, there has not been any research extracting clinical features in multiple time scales effectively.

\subsection{Interpretability of EMR Analysis}

On the one hand, the interpretability shown in most of the existing works mainly focuses on visit-level attention.
For example, some researches proposed RNN-based models with attention mechanisms to measure the relationships of different visits \cite{ma2017dipole,lee2018diagnosis}.

On the other hand, several researches have also explored the interpretability in the medical-feature-level.
Timeline \cite{bai2018interpretable} utilizes self-attention to generate clinical visit embedding, but can only identify disease code-level importance.
Some researches show the importance of features via adversarial attack, which is not an end-to-end framework \cite{sun2018identify}.
RETAIN \cite{choi2016retain} is more closely related to our work in terms of interpretability, which achieves feature-level interpretability by using attention mechanisms.
However, the prediction performance of RETAIN is limited \cite{ma2018health,ma2018risk}, due to the deficiency of effective high-level clinical feature extraction.
Existing studies still cannot capture the importance of biomarkers dynamically and meanwhile gain a performance boost in an end-to-end deep learning-based healthcare predictive model.

\section{Preliminary}

\subsection{A Motivating Example}
We take the health status prediction of end-stage renal disease (ESRD) patients as the motivating example. Currently, many people are suffered from ESRD in the world~\cite{tangri2011determining,isakova2011fibroblast}.
They face severe life threats and need lifelong treatments with periodic visits to the hospitals for multifarious tests (e.g., blood routine examination).
The whole procedure needs a dynamic patient health risk prediction to help patients recover smoothly and prevent the adverse outcome, based on the medical records collected along with the visits.
The core task of \mname is to learn the health status representation of the patient and perform the healthcare prediction.

\begin{table}[h]
    \centering
       \caption{Notations used in \mname}
    \label{tab:notations}
    \resizebox{\columnwidth}{!}{
    \begin{tabular}{l|l}
        \hline
        Notation & Definition \\
         \hline
         $y_{t}$ & Groundtruth of prediction target at $t$-th visit \\
         $\hat{y}_t$ & Prediction result at $t$-th visit \\
         \hline
         $\bm{r}_{t} \in \mathbb{R}^{N_{r}}$ & Multivariate visit record at $t$-th visit \\
         $\bm{u}_{t}^{r} \in \mathbb{R}^{|\bm{r}_{t}|}$ & Feature recalibration weight of $\bm{r}_{t}$\\
         $\widetilde{\bm{r}}_{t} \in \mathbb{R}^{|\bm{r}_{t}|}$ & Weighted input record \\
         \hline
         $\bm{c}_{t} \in \mathbb{R}^{K*N_{c}}$ & Extracted convolutional embedding of $\bm{r}_{t-L+1: t}$\\
         $\bm{u}_{t}^{c} \in \mathbb{R}^{|\bm{c}_{t}|}$ & Scale-adaptive recalibration weight of $\bm{c}_{t}$\\
         $\widetilde{\bm{c}}_{t} \in \mathbb{R}^{|\bm{c}_{t}|}$ & Weighted convolutional embedding \\
         \hline
         $\bm{v}_{t} \in \mathbb{R}^{N_{r}+K*N_{c}}$ & Visit embedding at $t$-th timestep\\
         $\bm{h}_{t} \in \mathbb{R}^{N_{h}}$ & Hidden state of GRU at $t$-th timestep\\
         \hline

         \hline
    \end{tabular}}
\end{table}

\subsection{Problem Formulation}

We assume that a patient overall visits clinic $T$ times, generating time-ordered EMR records denoted as $\bm{r}_{t}\in\mathbb{R}^{N_{r}}$ $(t = 1, 2,$ $\cdots$ $, T)$. Each EMR record contains $N_r$ features such as different lab test results. 
Thus the prediction problem in this paper can be formulated as, given $t$ historic EMR data of a patient, i.e., $(\bm{r}_{1}$, $\cdots$ , $\bm{r}_{t})$, how to predict the patient's healthcare status $y_t$ which is the probability of suffering from the specific risk (e.g., mortality risk, disease diagnosis, decompensation). The next section will detail our solution \mname.

\section{Methods}

\begin{figure}[]
  \centering
  \includegraphics[width=0.95\columnwidth]{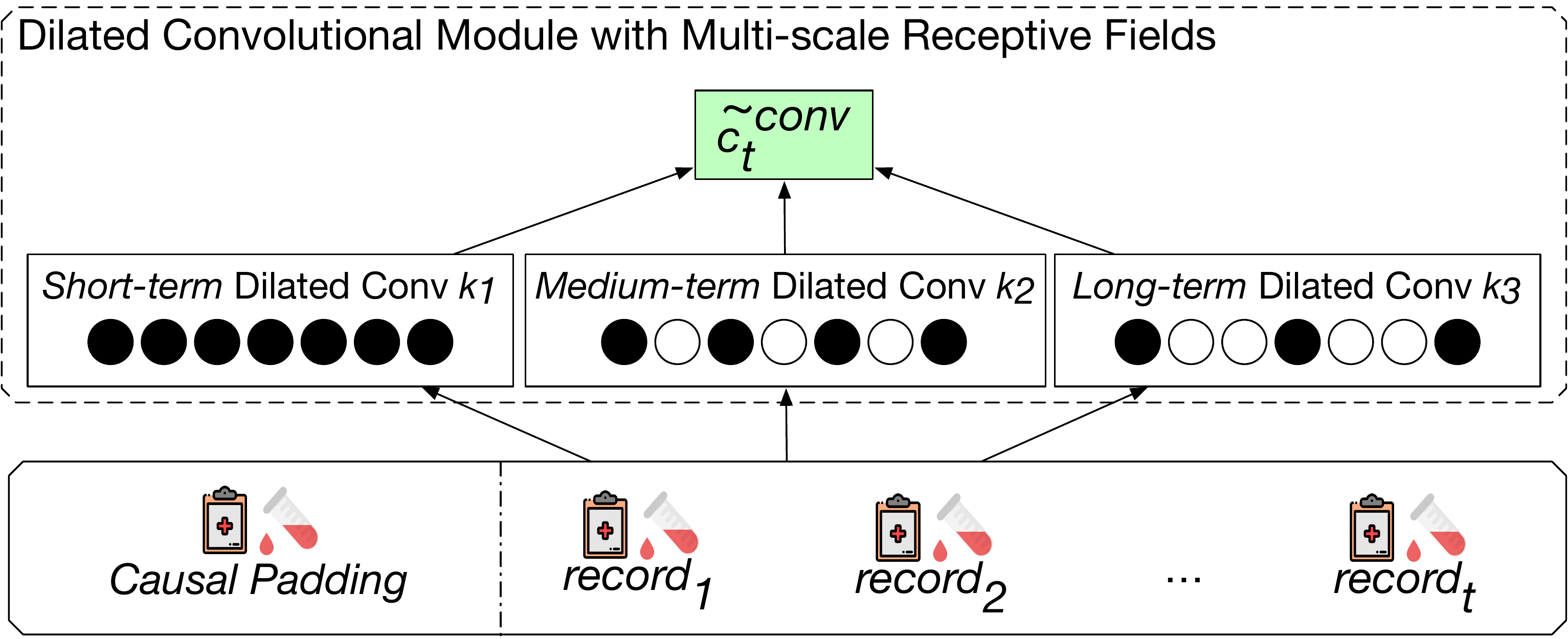}
  \caption{Capture the variation pattern of biomarkers in diverse time scales by dilated convolutional module with multi-scale receptive fields.}

  \label{fig:dilatedconv}
\end{figure}

\begin{figure}[]
  \centering
  \includegraphics[width=0.95\columnwidth]{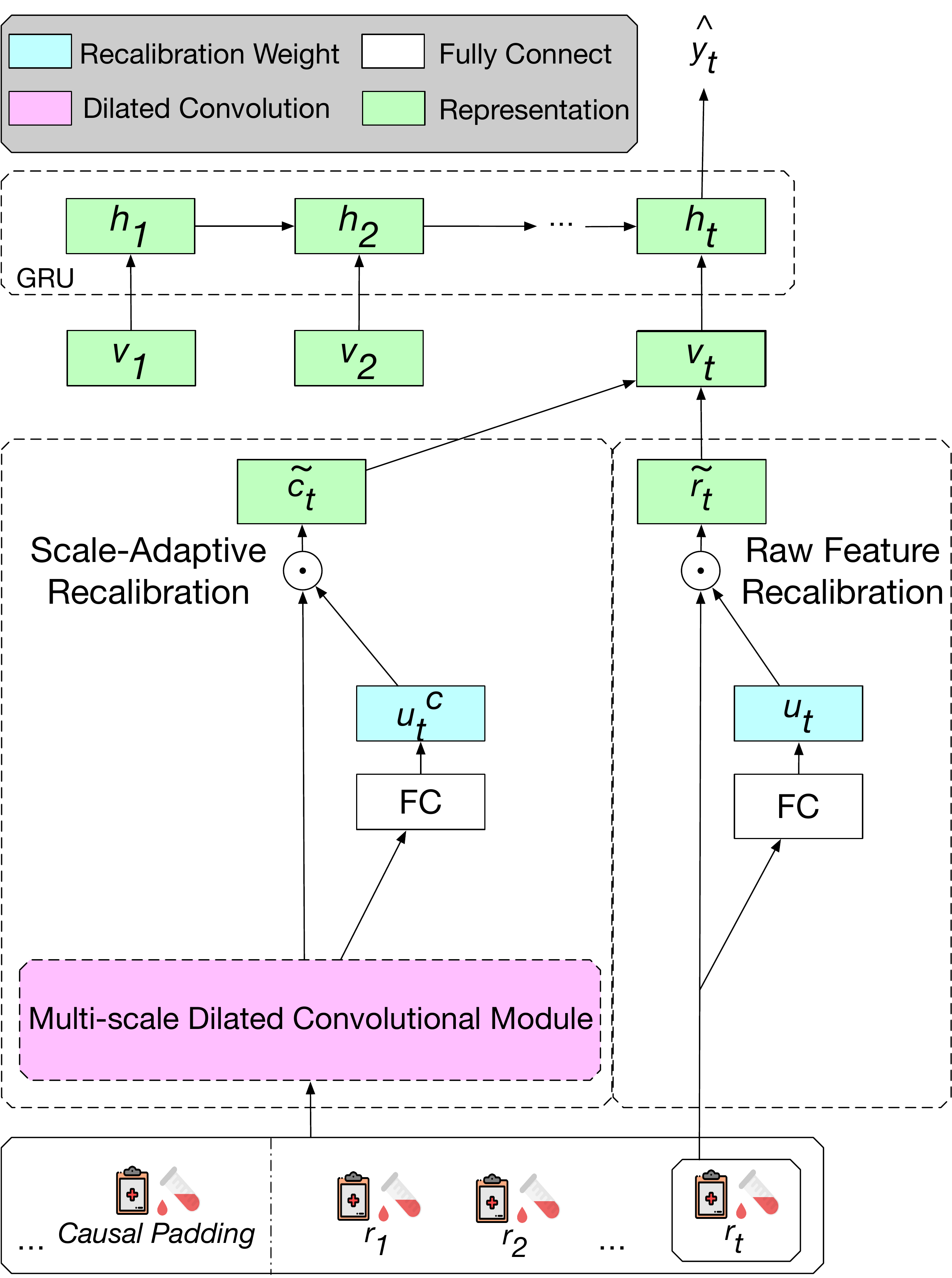}
  \caption{\mname Framework. Historical variation patterns of the biomarker in diverse time scales are extracted by a scale-adaptive dilated convolutional module. Both historical characteristics and the most recent record, which is attended by feature recalibration, are combined as clinical visit health status embedding. \mname can adaptively capture the predictive feature extracted in proper time scales and indicate the high-risk for the current health condition.}
  
  \label{fig:timeaware}
\end{figure}

Figure~\ref{fig:timeaware} shows the model structure of \mname, a Gated Recurrent Units (GRU) based architecture is used to embed the health status at each clinical visit and perform the healthcare prediction. 
The visit record sequence is embedded by the GRU to obtain the hidden state $\bm{h}_{t}$.
On the one hand, if the model depends on the latest record $\bm{r}_t$ alone, it would be overly sensitive to abnormal values of $\bm{r}_t$ brought by the missing data and noise of EMR, thus the prediction may lack robustness.
On the other, if the model only depends on the historical characteristics, the alertness of its prediction will be compromised.
In \mname, both of the historical characteristics $\bm{c}_t$ extracted by multi-scale dilated convolutional module and the $\bm{r}_t$ are taken into consideration to build the visit embedding $\bm{v}_t$.
Finally, we use $\bm{h}_{t}$ to predict $\widehat{y_t}$.
In summary, the novelty of \mname lies in the following two model components:
\begin{enumerate}
    \item As Figure~\ref{fig:dilatedconv} shows, we develop a dilated convolution~\cite{yu2015multi} with multi-scale receptive fields to capture the variation characteristics of biomarkers (addressing \textbf{$I_1$}).
    \item As illustrated in Figure~\ref{fig:timeaware}, we extend the squeeze and excitation block~\cite{hu2018squeeze} to dynamically capture the clinical features which strongly indicate the health risk (addressing \textbf{$I_2$}).
\end{enumerate}

\subsection{Multi-Scale Dilated Convolution}

One of our goals is to capture the dynamic variations of biomarkers over time and extract such local patterns as additional clinical features. 
But RNN module alone is difficult to achieve this.
The work~\cite{yu2015multi} demonstrated the effectiveness of $Dilated$ $Convolution$ to extract local patterns on images. 
Thus, \mname adopts a similar idea by adding a convolution filter before GRUs. 
But different to~\cite{yu2015multi}, we extend the dilated convolutional layers with different time spans (i.e., receptive fields), as depicted in Figure~\ref{fig:timeaware}. By doing so, \mname demonstrates the remarkable capability to capture both long-term trends and short-term abnormal variations of biomarkers simultaneously.
\begin{figure}[]
  \centering
  \includegraphics[width=0.95\columnwidth]{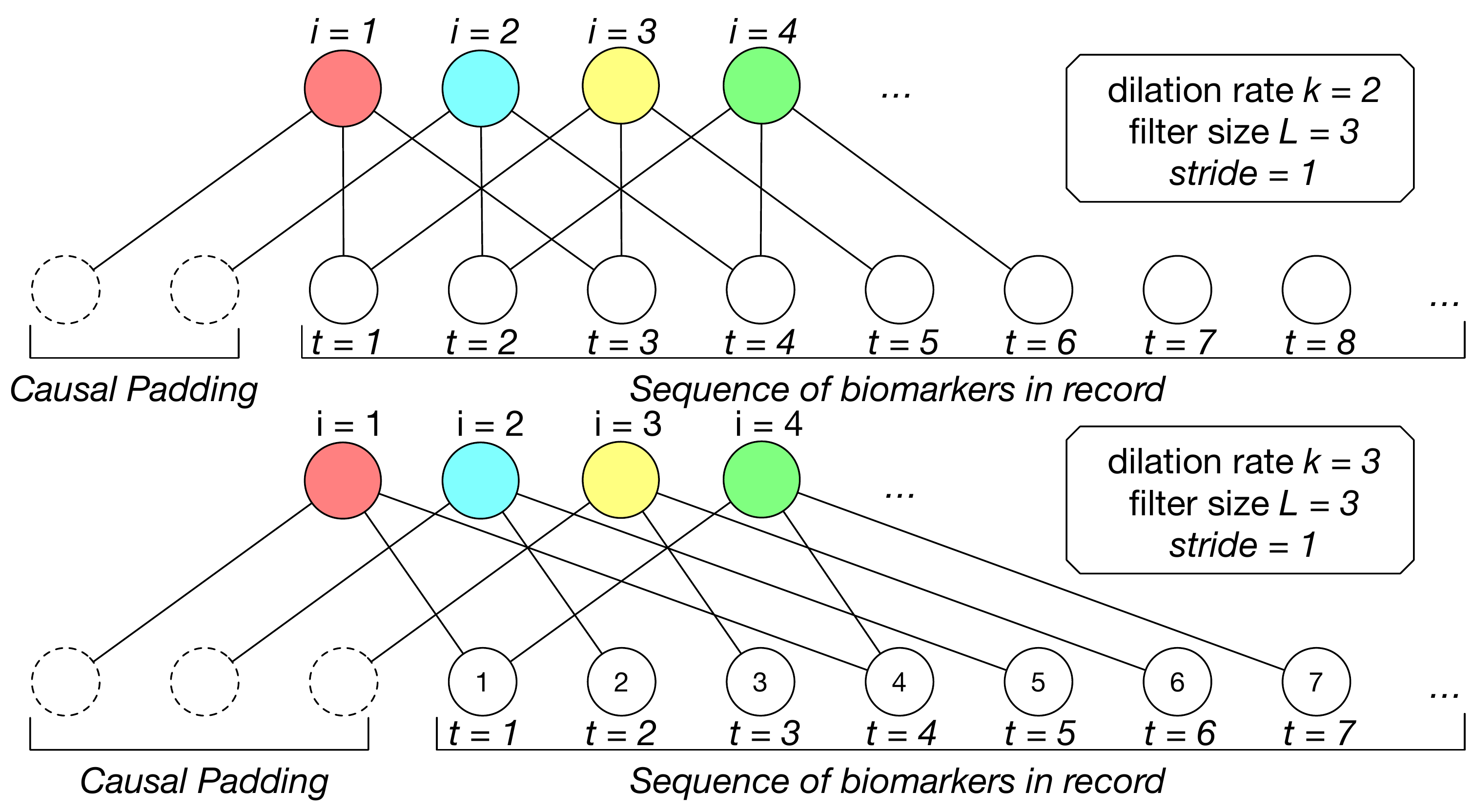}
  \caption{Dilated Convolutional Layers: { given the EMR sequence, dilation rate $k=1$ is a short-term standard convolution, and $k=2$ means medium-term convolution skipping one record per input, and $k=3$ means relative long-term convolution skipping two records}}
  \label{fig:dilated}
\end{figure}
Mathematically, the $Dilated$ $Convolution$ is a convolution applied to input with defined gaps:
\begin{equation}
d[i] = \sum_{l = 1}^{L} b[i+k\cdot l] \cdot z[l], 
\end{equation}
where 
$b$ is the input biomarkers of records, 
$d[i]$ is the output feature map, 
$z[l]$ denotes the convolutional filter of length $L$, 
and $k$ corresponds to the dilation rate. 
We use multiple filters to generate different filter maps, and the number of filter maps is $N_{c}$. 
We concatenate the multiple filter maps to get the final convolution output, denoted as ${conv}^{k}_{t}$.
Figure~\ref{fig:dilated} shows how dilated convolutional Layers with different dilation rates work.

In \mname, the dilated convolutional module is extended with multi-scale receptive fields consists of multiple parallel convolutional branches with the same filter size and stride but different dilation rates of $k_1$, $k_2$, $...$, $K$. 
For a given dilation rate, each layer takes multiple records of a time span, and the filters scan across the records to generate feature maps, which are concatenated to represents the long and short-term variation by
$\bm{c}_{t}= [{conv}^{k_1}_{t}; {conv}^{k_2}_{t};...; {conv}^{K}_{t}]$.
Moreover, to prevent the leakage of follow-up records, we utilize causal padding~\cite{yu2015multi}.

\subsection{Scale-Adaptive Clinical Feature Recalibration}

As we extract the sophisticated variation patterns of biomarkers, the multi-scale dilated convolutional module also introduces redundancy into the model inevitably.
Besides, some of the clinical features recorded in EMR have a high correlation with each other or even contribute little to the prediction target. 
It will reduce the interpretability and robustness of the learned representation if such redundant information is fed into the network.
To improve the adaptability of \mname in terms of feature utilization, we design the scale-adaptive clinical feature recalibration module based on $SEblock$ \cite{hu2018squeeze}.
This module is trained to model the nonlinear dependencies between clinical features explicitly. For a particular patient, it can selectively give more weights to the representative and predictive features but suppresses the unimportant features.
For patients with different disease conditions (e.g., suffering from chronic disease), the representative and predictive features in the corresponding time scale (e.g., the long-term dilated convolutional feature) would be enhanced.
Concretely, we design two fully-connected layers to learn the abstract weight representation and then re-scale it to the original dimension. 
\begin{equation}
\bm{u}^{c}_{t} = \sigma(\bm{U}_{c} ReLU (\bm{W}_{c} \bm{c}_t)),
\end{equation}
where $\bm{c}_{t}$ is the input of the abstraction operation;
parameter matrix $\bm{W}_{c}$$\in$$\mathbb{R}^{\frac{|\bm{c}_{t}|}{r}\times {|\bm{c}_{t}|}}$, $\bm{U}_{c} \in \mathbb{R}^{|\bm{c}_{t}| \times \frac{|\bm{c}_{t}|}{r}}$;
$r$ is the compress ratio that determines the abstraction degree of features;
$\bm{U}_{c}$ denotes the mapping matrix which rescales the input into $N_c$-dimensional;
$\sigma$ is the $Sigmoid$ activation function.
Then the learned weight $\bm{u}^{c}_{t}$ can be applied to the original features with an element-wise multiplication:
\begin{equation}
\widetilde{\bm{c}}_{t} = \bm{u}^{c}_{t} \odot \bm{c}_{t}
\end{equation}
The original input vector $\bm{c}_t$ is filtered to be sparser, and the redundancy of the network is reduced.
Such feature recalibration can be adjusted adaptively and dynamically through the visits according to the particular health condition.

Besides, in \mname, both the raw features and the features captured by the dilated convolutional layers are used to represent the current health status together via a recalibrated skip-connection.
The selectively enhanced predictive features can be treated as a precursor of health risk for the given patient.
\begin{equation}
    \bm{u}_{t}^{r} = \sigma(\bm{U}_{r} \cdot ReLU (\bm{W}_{r} \cdot \bm{r}_t)),
\end{equation}
\begin{equation}
\widetilde{\bm{r}_{t}} = \bm{u}_{t}^{r} \odot \bm{r}_{t}.
\end{equation}
The weighted raw features and the weighted convolutional features are concatenated together:
$\bm{v}_{t}=  [\widetilde{\bm{r}}_{t}; \widetilde{\bm{c}}_{t}]$.
The visit embedding $\bm{v}_{t}$ are fed into GRU to obtain hidden representations: 
$\bm{h}_t = GRU(\bm{h}_{t-1}, \bm{v}_t)$.
The attention mechanism can be easily adopted on the hidden representation here, but it is not the primary concern in this paper.
Finally, the healthcare prediction is obtained as:
\begin{equation}
    \widehat{y}_{t} = \sigma({\bm{W}_{y}\bm{h}_t+\bm{b}})
\end{equation}

\section{Experiment}

We conduct the decompensation prediction experiment on the MIMIC-III dataset and mortality prediction experiment on the ESRD (i.e., end-stage renal disease) dataset.
The effectiveness of dynamic feature recalibration is investigated by an overall observation.
In order to intuitively show the implication and prediction process of \mname, we also develop a simple visualization prototype.
The source code of \mname, statistics of datasets, case studies, and the visualization prototype are available at the GitHub repository\footnote{\url{https://github.com/Accountable-Machine-Intelligence/AdaCare}}.



\subsection{Data Preprocessing and Prediction Tasks}
\begin{itemize}[leftmargin=*]

\item
\textbf{MIMIC-III Dataset \footnote{https://mimic.physionet.org}.} We use ICU data from the publicly available Medical Information Mart for Intensive Care (MIMIC-III) database~\cite{johnson2016mimic}.
We perform the detection of physiologically decompensating patients, which is formulated as a binary classification problem based on patients' clinical events produced during ICU stays \cite{harutyunyan2017multitask}. 
Physiologic decompensation is formulated as a problem of predicting if a patient would die within the next 24 hours by continuously monitoring the patient within fixed time-windows.
Decompensation labels were curated based on the occurrence of the patient's date of death (DOD) within the next 24 hours, and only about 4.2\% of samples are positive in the benchmark.
There are 1,203 patients (about 2.89\%) with overlong sequences (i.e., $>400$).
Without loss of fairness, we truncate the length of samples to a reasonable limit (i.e., 400).
Eventually, a cohort of 41, 602 unique patients with a total of 3,431,622 samples (i.e., records) is used in our dataset. 
We fix a test set of 15\% of patients and divide the rest of the dataset into the training set and validation set with a proportion of 85\%:15\%.
We resample the test set 1000 times using the bootstrap method \cite{harutyunyan2017multitask} and calculate the standard deviation of the results.

\item 
\textbf{ESRD Dataset.} We perform the mortality risk prediction on a real-word end-stage renal disease dataset.
In this study, all end-stage renal disease patients who received therapy from January 1, 2006, to March 1, 2018, in a real-world hospital are included to form this dataset. 
We select the features that are observed in more than 60\% patients' records. 
For missing values, we fill the missing front cells with the data backward to prevent the leakage of future information. 
If the backward record of a patient is missing, we impute it with the first front observed record of the patient.
The cleaned dataset consists of 656 patients with static baseline information and 13,091 dynamic records. 
There are 1196 records with positive labels (i.e., died within 12 months) and 10,804 records with negative labels.
We evaluate the models with a 10-fold cross-validation strategy and report the average performance, similar to \cite{ma2018health}.


\end{itemize}

Similar to related researches, we assess performance using
area under the precision-recall curve (AUPRC) \cite{keilwagen2014area},
the minimum of precision and sensitivity Min(Se,P+),
and area under the receiver operating characteristic curve (AUROC) \cite{hanley1982meaning}. The Min(Se,P+) is calculated as the maximum of min(sensitivity, precision) on the precision-recall curve.



\begin{table*}[]
  \centering
  \caption{Results of Health Risk Prediction. Values in the parentheses denote standard deviations.}
  \label{tab:result}

\begin{tabular}{lcccccc}
\hline
 & \multicolumn{3}{c}{Mortality Prediction on ESRD} & \multicolumn{3}{c}{Decompensation Prediction of MIMIC} \\

   & AUPRC & min(Se, P+) & AUROC  & AUPRC & min(Se, P+) & AUROC \\ 
\hline
GRU  & 27.14\% (.025) & 31.66\% (.030) & 80.66\% (.013) & 27.84\% (.003) & 32.60\% (.004) & 89.83\% (.003)  \\
 
 
RETAIN  & 26.18\% (.021) & 29.98\% (.033) & 79.25\% (.027) & 25.97\% (.004) & 29.00\% (.005) & 87.64\% (.002)\\

T-LSTM   & 27.84\% (.019) & 33.37\% (.028) & 81.13\% (.021)  & 26.11\% (.003) & 31.86\% (.004) & 89.44\% (.002) \\

SAnD$_{*}$  & 26.31\% (.033) & 29.94\% (.037) & 79.54\% (.032)  & 25.24\% (.003) & 28.99\% (.004) & 88.25\% (.003) \\

 \hline
$\mname_{-,\sigma}$   & 27.66\% (.025) & 30.55\% (.039) & 79.64\% (.028)  & 28.11\% (.002) & 32.71\% (.003) & 89.77\% (.003)\\
$\mname_{c,-}$   & 30.77\% (.021) & 33.45\% (.022) & 81.22\% (.012)  & 28.37\% (.003) & 33.10\% (.003) & 89.81\% (.002)\\


$\mname_{c,\varsigma}$   & 30.98\% (.025) & 33.31\% (.033) & 80.61\% (.019) & 28.95\% (.004) & 34.23\% (.004) & 89.93\% (.002) \\

\hline

$\mname_{c,\sigma}$   & \textbf{31.79\% (.020)} & \textbf{34.46\% (.030)} & \textbf{81.51\% (.017)} & \textbf{30.37\% (.004)} & \textbf{34.29\% (.004)} & \textbf{90.04\% (.003)}\\



\hline
\end{tabular}
\end{table*}










\subsection{Implementation Details and Baselines}

Several models share part of the similar insights with \mname to learn the representation of patient status, some of which are taken as baseline approaches and listed as follows. We conduct a grid search over hyper-parameters space for the models.

\begin{itemize}[leftmargin=*]
\item \textbf{GRU} is the standard Gated Recurrent Unit network. 

\item \textbf{RETAIN} \cite{choi2016retain} utilizes a two-level neural attention mechanism to detect influential visits and significant variables, which provide interpretability.

\item \textbf{T-LSTM} \cite{baytas2017patient} handles irregular time intervals by enabling time decay. We modify it into a supervised learning model.

\item \textbf{SAnD$_{*}$} \cite{song2018attend} models clinical time-series data solely based on self-attention. We re-implement SAnD by using $r_{t-k+1:t}$ to build input embedding at the measurement position $t$ (i.e., causal padding \cite{van2016wavenet}), instead of the one proposed in the original paper $r_{t:t+k-1}$, to avoid the violation of causality. 
The kernel size of convolutional embedding is set to 1.
\end{itemize}

We also compare \mname with the variants of our approaches.
Subscript $c$ in Table \ref{tab:result} denotes the multi-scale dilated convolution.
Subscripts $\sigma$ and $\varsigma$ denote the raw feature recalibration module learned with activation function $sigmoid$ and $sparsemax$ respectively.

For \mname, we set the hidden units of \mname to 64 for the ESRD dataset and 128 for MIMIC-III dataset. We use 64 filters for convolutional layers, the kernel size is set to 2, and the dilation rate is set to 1,2,3/1,3,5 for ESRD/MIMIC dataset, respectively. 
For the feature recalibration block, we set the compression ratio to 2/4 for ESRD/MIMIC dataset, respectively. We also utilize the dropout strategy (the dropout rate is 0.5) between the RNN layer and the final output layer for all approaches. We utilize Adam optimizer \cite{kingma2014adam} with the mini-batch of 128 patients, and the learning rate is set to $1e-3$.
The training is done on a machine equipped with CPU: Intel Xeon E5-2630, 256GB RAM, and GPU: Nvidia Titan V. We implement \mname with Pytorch 1.1.0. 

\subsection{Results of Healthcare Prediction}

Table \ref{tab:result} shows that the performance of all approaches on two datasets: MIMIC-III and the ESRD dataset. 
\mname outperforms all baseline models across both datasets in all evaluation metrics, especially AUPRC, which is the most informative and the primary evaluation metric when dealing with a highly imbalanced and skewed dataset \cite{davis2006relationship,choi2018mime} like the real-world EMR data.
Compared to the best baseline model, \mname achieves relative improvements of 14.2\% and 9.1\% for AUPRC in ESRD and MIMIC dataset respectively. Although RETAIN can provide interpretability, its performance is worse than the basic GRU model on both two datasets, 
which is consistent with the results reported in \cite{ma2018risk}.

$\mname_{c,-}$ (i.e., with multi-scale dilated convolution) outperforms the baselines mentioned above.
It confirms our assumption that extracting the historical variation pattern of the biomarker in different time scales can depict the health status more comprehensively.
$\mname_{c,\sigma}$ (i.e., multi-scale dilated convolution and feature recalibration with $sigmoid$ activation function) outperforms the baseline approaches including $\mname_{c,-}$.
It suggests that the feature recalibration module can enhance the predictive feature to build the representation effectively, and improve the performance.

To further verify the effectiveness of the model when clearly providing the most high-risk clinical feature in latest visit for the physician,
we also test the performance of $\mname_{c,\varsigma}$, 
which utilizes $sparsemax$ as the activation function of the raw feature recalibration.
Such recalibration enhances only a few most predictive features and suppresses most of the features.
When performing the mortality prediction on the ESRD dataset, the performance of $\mname_{c,\varsigma}$ is slightly worse than $\mname_{c,\sigma}$, and consistently better than most of the comparative approaches.
The result indicates that the $\mname_{c,\varsigma}$ is still reliable when performing health prediction on the ESRD dataset.

\subsection{Interpretability and Implications}

\begin{figure}[]
  \centering
  \includegraphics[width=0.95\columnwidth]{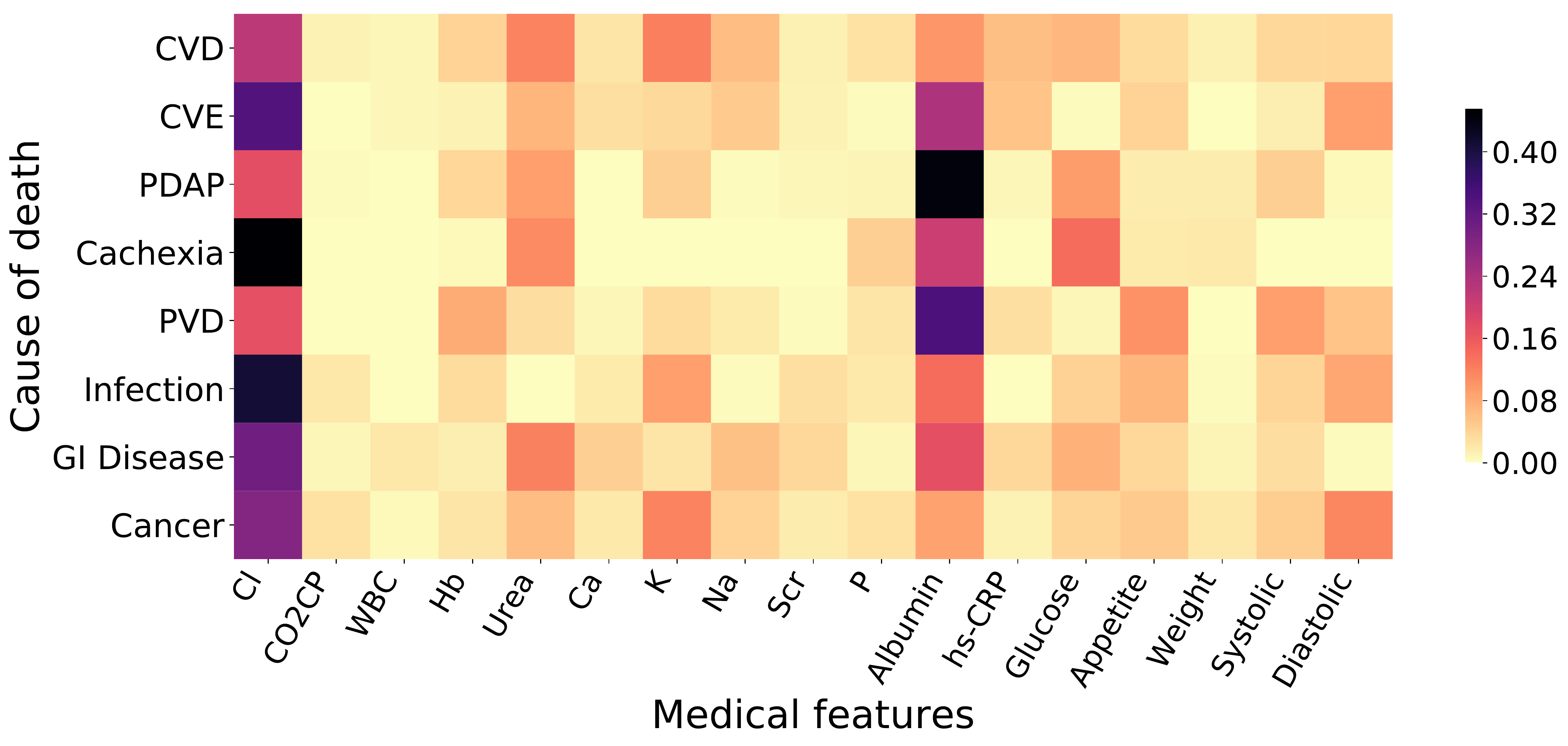}
  \caption{Importance of biomarkers differentiated by dynamic feature recalibration. 
  }
  \label{fig:rawattention}
\end{figure}

The case study of a specific sample is usually used to verify the interpretability in the EMR analysis researches,
but it is still not convincing enough due to the contingency of case studies.
In order to quantitatively identify the reasonability of feature recalibration from an overall perspective,
we calculate the average importance weights of biomarkers on different causes of death on ESRD validation sets.
The feature-death reason importance are visualized on Figure~ \ref{fig:rawattention}.
Some of the essential medical knowledge learned by \mname are summarized as follows:

\begin{itemize}
    \item Serum albumin is strongly related to adverse outcomes of ESRD patients, especially for the Peritoneal Dialysis-Associated Peritonitis (PDAP). 
    This is consistent with the medical researches \cite{blake1993serum,spiegel1993serum,cheng2008relationship,meijers2008review}, which figure out that the abnormal value and decreasing trend of serum albumin usually indicate that the patient may suffer from inflammation and fluid overload.
    
   \item Urea is related to gastrointestinal (Gl) disease and cachexia. According to medical research\cite{honda2006serum}, the abnormal value and decreasing trend of Urea usually indicate that the patient may suffer from low protein intake and malnutrition.
   
     \item Serum chlorine (Cl) is strongly related to adverse outcomes of ESRD patients, especially for the cachexia and infection. 
    According to medical experience, the serum chlorine level reflects the renal function of the patient to some extent.
    However, the relationship between infection, cachexia, and serum chlorine has not been fully explored by existing medical researches.
    This noteworthy medical finding has already raised the interest of medical experts.
   

\end{itemize}

We conduct application-grounded evaluation \cite{doshi2017towards} by inviting 12 experienced medical practitioners (with 5-15 years practicing time) from nephrology departments of 5 different hospitals, to evaluate the agreement degree of interpretability generated by \mname. 
The interpretability provided by \mname is highly consistent with the practice experience of human experts.
Some of the extracted medical knowledge has already been introduced as the ESRD management aid by physicians.
More details about the experiment are described at our GitHub repository.

\section{Conclusion}

In this paper, \mname is proposed to learn the clinical health status representation.
Specifically, we utilize the dilated convolutions with multi-scale receptive fields to capture the long and short-term historical variation of biomarkers.
\mname models the nonlinear dependencies of features by extending SE-block.
Such a feature re-calibration process selectively enhances predictive features extracted in proper time scales and the most high-risk factors in the latest visit.
It builds effective health status representation
and provide reasonable interpretability.
Experiment results on MIMIC-III dataset and ESRD dataset show that \mname outperforms the baseline approaches with powerful interpretability.
Medical knowledge learned by \mname has been positively confirmed by human medical experts and related medical literature.

\section{ Acknowledgments}
This work is supported by National Science and Technology Major Project (No. 2018ZX10201002) and the fund of Peking University Health Science Center (BMU20160584). WR is supported by ORCA PRF Project (EP/R026173/1).

\bibliographystyle{aaai}
\bibliography{3070_ref.bib}

\end{document}